\documentclass[conference]{IEEEtran}

\usepackage{latexsym}
\usepackage{graphicx}
\usepackage{amsmath}
\usepackage{url}
\usepackage{cite}
\usepackage{balance}

\newcommand{\ourMethod}{WCT-VEC}  
\newcommand{\bve}{UMFS-WE}
\newcommand{\wts}{{\sc Comp2Sense}} 
\newcommand{\lext}{Lesk$_\text{ext}$}
\newcommand{\mfsTask}{MFS detection}
\newcommand{\Random}{Random} 
\newcommand{\Oracle}{Supervised MFS} 
\newcommand{\oracle}{supervised MFS} 
\newcommand{\relatives}{companions}  
\newcommand{\Relatives}{Companions}  
\newcommand{\MFT}{most frequent translation}

\begin{document}

\title{You Shall Know the Most Frequent Sense\\ by the Company it Keeps}

\author{
  \IEEEauthorblockN{Bradley Hauer, Yixing Luan, and Grzegorz Kondrak}
  \IEEEauthorblockA{
    Department of Computing Science\\
    University of Alberta, Edmonton, Canada\\
    {\tt {\{bmhauer,yixing1,gkondrak\}}@ualberta.ca}
  }
}

\maketitle

\begin{abstract}
Identification of the most frequent sense of a polysemous word
is an important semantic task.
We introduce two concepts that can benefit {\mfsTask}:
\emph{\relatives}, 
which are the most frequently co-occurring words,
and the {\em most frequent translation} in a bitext.
We present two novel methods
that incorporate these new concepts,
and show that they
advance the state of the art on {\mfsTask}.
\end{abstract}

\IEEEpeerreviewmaketitle

\section{Introduction}
\label{intro}

{\mfsTask} is the task of identifying the most frequent sense 
of a polysemous word.
The task must be defined with respect to a particular body of text.
For example, 
one would expect {\em bank} to refer to a river bank in a geographic text,
but the ``repository'' sense is the most frequent in general English.

{\mfsTask} is important in 
word sense disambiguation (WSD),
the area of research concerned with 
determining the meaning
of words in context.
WSD systems can use the MFS as a back-off method,
or as an additional source of information.
An MFS-based WSD system that
classifies each word token according to its MFS
is a strong WSD baseline that 
typically outperforms unsupervised WSD systems,
and approaches the accuracy of supervised systems \cite{raganato2017}.
{\mfsTask} is also an interesting task itself,
which could be applied, for example, 
to provide the predominant sense of a word 
in an interactive dictionary look-up.

{\mfsTask} is an unsupervised classification problem.
Although
sense frequency information
can be approximated from a large sense-annotated corpus, 
such resources are expensive to create,
as hundreds of thousands of word tokens need to be manually disambiguated.
On the other hand,
\mfsTask{} systems require only unannotated text corpora,
and so can be more easily applied to different domains and languages.

\cite{mohammad2006emnlp} generalize 
 the famous observation of \cite{firth1957}
as
``you shall know a sense by the company it keeps.''
We propose to apply this intuition to {\mfsTask},
as stated in this paper's title.
Specifically,
our hypothesis is that 
the most frequent sense of a given target word 
can be determined by referring to the set of words
that most frequently co-occur with it,
which we refer to as the word's {\em companions}.
In addition, 
following the observation 
that different senses of a word may translate differently,
we propose that leveraging frequent word translations from
a bitext can improve accuracy of a {\mfsTask} system.

In order to test our hypothesis,
we develop two  
knowledge-based methods for \mfsTask{}.
The first method selects
the sense which is most closely related to its companions,
according to a WordNet-based sense similarity measure.
The second method 
constructs a series of vectors
for words, senses, companions, and most frequent translations,
and selects the MFS on the basis of the cosine similarities between them.
The principal contributions of this work are
the introduction and application of the concepts of companions 
and most frequent translations to the task of {\mfsTask}.

We conduct an extensive evaluation of the proposed methods,
which includes a series of intrinsic, extrinsic, and ablation experiments
on standard datasets,
as well as error analysis.
The experimental results establish a new state of the art for {\mfsTask}.
In order to 
facilitate replication 
and encourage further work on \mfsTask{},
we publish our {\mfsTask} results
for all words covered by WordNet in a large, unannotated 
bitext.\footnote{\url{https://webdocs.cs.ualberta.ca/~kondrak/}}

\section{Related Work}
\label{relwork}

\cite{buitelaar2001}
lay the groundwork for \mfsTask{}
by 
analyzing the relevance of GermaNet
synsets to specific domains.
\cite{koeling2005} build upon this,
showing that 
WSD performance can be improved by
performing \mfsTask{}
on a corpus of the same domain as the testing data.
\cite{mohammad2006} present an \mfsTask{} method 
based on a published thesaurus,
which they use to
induce a coarse-grained sense inventory.
This separates their method from other related work,
which typically uses WordNet 
as the \emph{de facto} sense inventory for WSD and \mfsTask{}.

\cite{mccarthy2004} present a method for \mfsTask{}
based on a thesaurus constructed from a parsed corpus.
This thesaurus is used to induce a word similarity function,
which they use to assess the prevalence of each sense of a given target word.
They perform both intrinsic and extrinsic evaluations;
we compare to their reported results 
to the extent their experimental setup allows.
This method was subsequently applied 
to a WSD shared task \cite{mccarthy2004se3},
and to the identification of infrequent word senses \cite{mccarthy2004coling}.
An extended analysis of the method was presented by \cite{mccarthy2007}.
\cite{iida2008} adapt this method
to Japanese \mfsTask{} using only the glosses of words,
excluding the use of semantic networks such as WordNet.

\cite{bhingardive2015} present the first \mfsTask{} method
based on automatically learned vector word embeddings.
They test their method on English and a private Hindi dataset.
This is the most recent work we are aware of 
which considers the exact same task as we do, 
in the same setting;
given its recency relative to other such works,
we consider this to be the state-of-the-art for \mfsTask{}.
We re-implement this method,
and compare to it directly in our experimental evaluation.

\mfsTask{} is related to, but distinct from,
the task of \emph{sense distribution learning} (SDL),
in which the goal is to predict the frequency distribution
of the senses of a given target word.
Prior work on SDL includes 
\cite{chan2005ijcai}, \cite{lau2014}, 
\cite{bennett2016}, and \cite{pasini2018},
In principle, a SDL system can be applied to {\mfsTask}
by simply returning the sense 
with the highest probability.
We compare our results to 
the EnDi method of \cite{pasini2018},
which is the current state of the art in SDL.

As part of one of our methods,
we induce vector representations not only of words,
but also of individual senses.
Our method differs from recent prior work on
constructing embeddings using sense information
\cite{rothe2015,camacho2016,pilehvar2016,suster2016};
instead, we build upon the
methods of \cite{chen2014} and \cite{bhingardive2015}.
The resulting vectors are 
easier to create with fewer resources,
more interpretable,
and easier to extend and compare to other types of vectors
such as those we construct in Sections \ref{relvec} and \ref{mftvec}.
We are particularly interested in maintaining the ability
to perform semantic comparisons across different languages,
as demonstrated by prior work \cite{mikolov2013},
motivating our decision to work with vectors known
to have this property.

Our methods leverage
cross-lingual information 
and contextually related words.
These concepts have previously been used
to improve WSD
-- \cite{yarowsky1995,ng2003,navigli2009,apidianaki2015}, and others --
but our usage of these concepts for \mfsTask{} is novel.

\section{\wts}
\label{methods2}

Given a target word,
our first \mfsTask{} method 
uses a set of words known as its \emph{\relatives{}}
to determine its MFS.
The method is based on a sense-similarity function
which makes use of WordNet's hierarchical semantic network.
Since the method relates the \relatives{} 
of the target word
to its senses,
we name it \wts{}.

For each word $w$ in a given corpus,
we define the {\relatives} of $w$
to be the $k$ content words, other than $w$ itself,
which most frequently occur in sentences containing $w$.
The variable $k$ is a tunable parameter.
Building on prior work \cite{mohammad2006},
the \relatives{} of a word are defined in an entirely relation-free way,
requiring no external resources or pre-processing to extract
(this distinguishes our method from prior work,
e.g. \cite{pantel2002}, \cite{mccarthy2004}).
We experimented with more sophisticated methods for selecting \relatives{},
such as taking the $k$ words with the highest pointwise mutual information
with the target word,
but in our development experiments,
taking the $k$ most frequently co-occurring words
gave substantially better results.

Our method uses
the WordToSet\footnote{Formally, \textsc{WordNet::SenseRelate::WordToSet}}
package \cite{pedersen2005}
as a subroutine.
WordToSet takes as input a word $w$,
and a set of words $X$,
and compares the senses of $w$
to the senses of each word in $X$,
returning the sense of $w$
which is found to be most closely related to the words in $X$.
The default similarity function for performing sense-to-sense comparison
is $jcn$ \cite{jiang1997}.
Given a pair of senses $s$ and $s'$,
$jcn(s,s')$ is a real number,
such that the more closely related the given senses are
with respect to the WordNet sense hierarchy,
the higher the returned value.
This algorithm was developed for WSD,
and variants of it are still used as strong knowledge-based 
WSD baselines~\cite{raganato2017}.
We apply WordToSet to \mfsTask{} for the first time.

For a word $w$, 
let $C_w$ be the set containing the \relatives{} of $w$, and
let $S_w$ be the set containing the senses of $w$.
Note that $C_w$ is a set of words,
while $S_w$ is a set of senses.
To identify the MFS of a target word $w$,
\wts{}
uses WordToSet
to assign a score to each $s \in S_w$ as follows:
\[
score(s) = \displaystyle\sum\limits_{c \in C_w}
           \displaystyle\max\limits_{s' \in S_c}
           jcn(s, s')
\]
The sense with the highest such score is returned.

\section{\ourMethod{}}
\label{methods}

\begin{figure}[t]
\centering
\includegraphics[width=\linewidth]{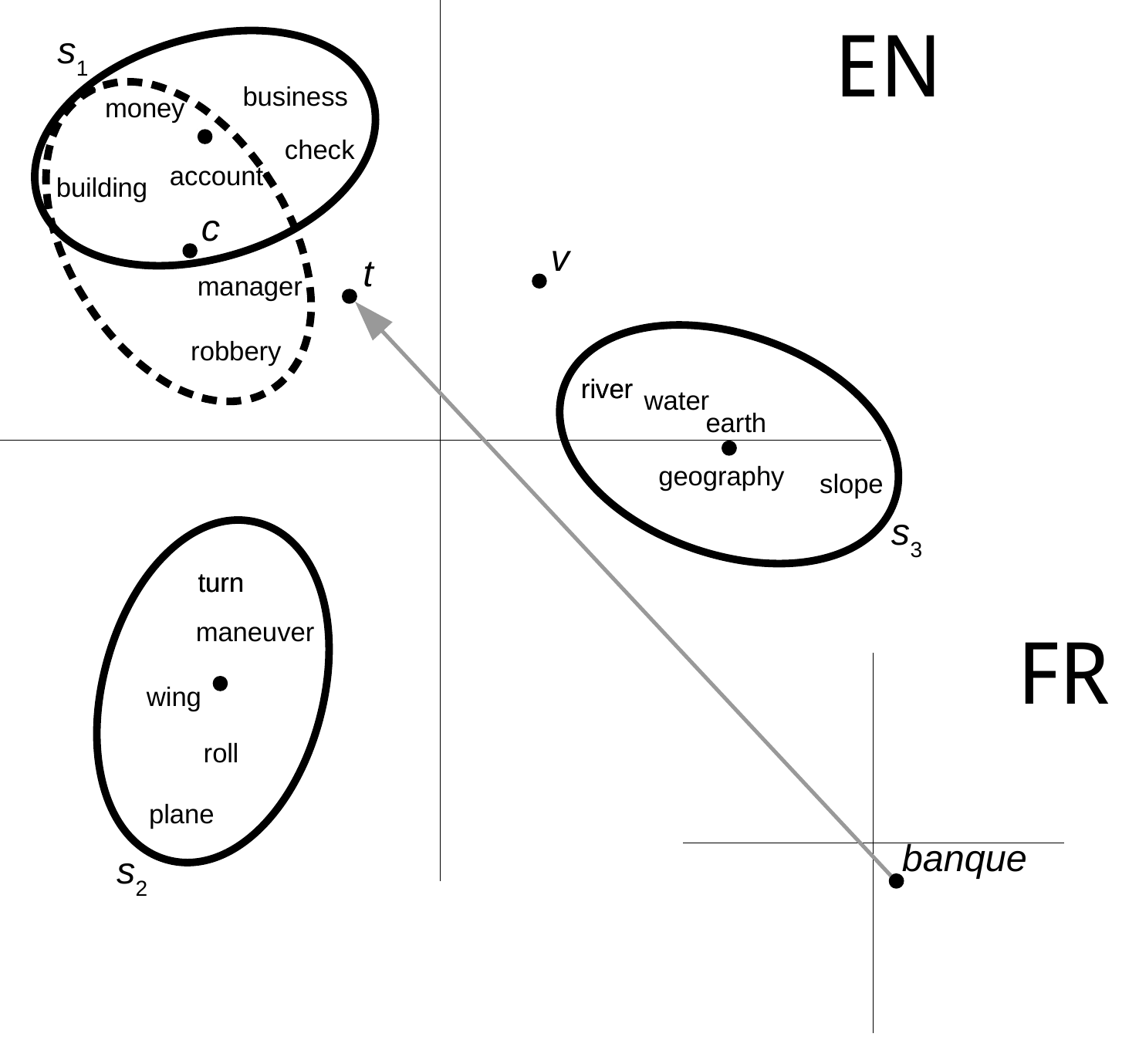}
\caption{A simplified illustration of the vectors used
to identify the MFS of the word ``bank'':
the sense vectors ($s$) with the associated keywords,
as well as the word ($v$), companions ($c$), and MFT ($t$) vectors,
with the latter translated from the French vector space (FR).}
\label{2d-diagram}
\end{figure}

Our second \mfsTask{} method is based on
vector embeddings of words,
which are constructed such that
cosine similarity of vectors 
approximates a measurement of semantic similarity.
The method amalgamates these word vectors
to create a \emph{sense vector} for each sense of the target word,
and compares each to three vectors
which represent 
the target word itself 
(Section~\ref{word2vec}),
its {\relatives}
(Section~\ref{relvec}),
and its most frequent translation in another language
(Section~\ref{mftvec}),
respectively.
These three vectors,
which depend only on the target word,
collectively represent the MFS,
and so the sense whose vector is closest
to these three vectors is taken to be the MFS
(Figure \ref{2d-diagram}).
We call this method \ourMethod{},
where WCT is an abbreviation of ``Word, \Relatives{}, and Translation''.

\subsection{Word Vectors}
\label{word2vec}

\ourMethod{} begins from word embeddings,
low-dimensional real-valued vector representations 
of each word in the vocabulary,
which can be compared using cosine similarities
as described above.
We create such vectors using \textsc{word2vec} \cite{word2vec}, 
a well-known software package
which learns vector embeddings from unlabelled monolingual data
using a simple neural model.
We denote the embedding of a word $w$ as $\mathbf{v}_w$.

We adopt the assumption,
supported by the work of \cite{arora2016},
that 
a vector representing a polysemous word
is a composition of vectors
representing each of its senses,
with more frequent senses having greater influence on the word vector.
Thus, the more frequent a sense of a word is,
the closer its sense vector should be to the vector of the word.

\subsection{Sense Vectors}
\label{sense2vec}

Since our approach is unsupervised,
we cannot train sense vectors directly using sense annotated data,
as done by, for example, \cite{mancini2017}.
Following prior work \cite{bhingardive2015} \cite{chen2014},
we instead approximate them  
by identifying {\em keywords} for each sense,
and taking the average of their vectors.
The use of external resources, such as WordNet, 
distinguishes the knowledge-based paradigm form the supervised paradigm
\cite{navigli2018}.

We differ from previous work in how we identify the sense keywords.
\cite{chen2014} select keywords 
from the sense gloss whose vectors are similar to the word vector,
using a threshold of cosine similarity.
\cite{bhingardive2015} add the synsets that contain:
(a) the synonyms, hypernyms, and hyponyms of the sense; 
(b) the content words from the glosses and usage examples of the sense; 
and (c) the keywords found using the above semantic relationships.
We extended this method
by expanding our keyword sets to also include
meronyms, holonyms, entailments, causes,
and similar words (as encoded in WordNet).

\subsection{\Relatives{} Vector}
\label{relvec}

As with our {\wts} method, 
\ourMethod{} also leverages the notion of {\relatives}.
Rather than comparing the senses of a target word $w$
to each sense of each of its \relatives{},
\ourMethod{} represents the \relatives{} of $w$ with a single vector,
which is the average of the vectors of the \relatives{} of $w$.
We call this vector the \emph{\relatives{} vector} of $w$,
and denote it $\mathbf{c}_w$.
Following our hypothesis that the \relatives{} of a word
are most closely related to its MFS,
we expect $\mathbf{c}_w$  
to have higher cosine similarity with the vector of the MFS of $w$
than with the vector of any other sense of $w$.

\subsection{Most Frequent Translation Vector}
\label{mftvec}

Different senses of the same word 
may translate as different words in other languages \cite{resnik1997,ng2003}.
We leverage this fact to help identify the most frequent sense of a word
by identifying its \emph{most frequent translation (MFT)} 
in a sentence-aligned bilingual corpus (bitext).
We expect the most frequent translation of a word 
to be a translation of its most frequent sense.

Without loss of generality,
let the bitext represent English and French.
We word-align the bitext,
and define the MFT of each English word to be
the French word with which the English word is most frequently aligned.
After computing vector representations for the French vocabulary
using the French side of the bitext,
we proceed to 
learn a cross-lingual linear transformation
using the method of \cite{mikolov2013}.
This method takes a set $n$ of English--French translation pairs, $(e_i, f_i)$,
represented by their word embeddings 
in the English and French vector spaces respectively.
It then uses stochastic gradient descent to learn 
a \emph{translation matrix} $T$, by minimizing the objective function
\[
\displaystyle\sum\limits_{i=1}^n || T \cdot f_i \;-\; e_i ||^2
\]
This allows us to map the French word vectors
into the English vector space
in a way that preserves the semantic properties of the word vectors.
Different from \cite{mikolov2013},
rather than obtaining translation pairs from Google Translate,
we obtain training pairs from our word-aligned bitext,
using the most frequent translations 
of the $5000$ most frequent English words as training data,
as was previously demonstrated to be effective by \cite{hauer2017}.

We obtain the {\MFT} vector of $w$, denoted $\mathbf{t}_w$,
by computing $\mathbf{t}_w = T \cdot \mathbf{f}$,
where $\mathbf{f}$ is the embedding of the MFT of $w$ 
in the French vector space.
Thus, $\mathbf{t}_w$ represents the MFT of $w$
in a way that allows semantic comparison 
to vectors in the English vector space.

\subsection{Identifying the MFS}
\label{combine}

In the previous sections,
we have identified three intuitive properties of the MFS:
(1) the vector of the MFS should be closest to the vector of the target word
(Section \ref{word2vec});
(2) the MFS should be most closely related to the \relatives{} of the word
(Section \ref{relvec});
and (3) the MFT of the word should be a translation of its MFS
(Section \ref{mftvec}).
We have shown how we construct vector representations
of each sense of the target word,
as well as vectors 
which model these three sources of information regarding the target word itself.
These vectors admit efficient semantic comparison 
through computation of cosine similarities \cite{mikolov2013},
which we use to identify the MFS.

Given a target word type $w$ with 
word vector $\mathbf{v}_w$,
{\relatives} vector $\mathbf{c}_w$,
and  
MFT vector $\mathbf{t}_w$, 
as well as 
a set of senses $S$ 
with each $s \in S$ having an associated sense vector $\mathbf{s}$ 
(Section \ref{sense2vec}),
\ourMethod{} identifies the MFS of $w$ as follows:
\[ MFS(w) = \arg\max_{s \in S} \{\; 
     \chi_1 \cos(\mathbf{s}, \mathbf{v}_w) \]
\[ \quad\quad \;+\; \chi_2 \cos(\mathbf{s}, \mathbf{c}_w) 
   \;+\; \chi_3 \cos(\mathbf{s}, \mathbf{t}_w) \;\} \]
where the $\chi_i$ are tunable non-negative parameters 
which sum to 1.
Figure \ref{2d-diagram} illustrates a simplified example
of the vectors \ourMethod{} uses to identify the MFS of a word.

\section{Experiments}
\label{eval}

In this section,
we describe both intrinsic and extrinsic evaluation experiments,
comparing
our approach against previous work
on standard datasets.
In addition, we perform ablation experiments and error analysis.

\subsection{Experimental Setup}
\label{setup}

All sense-annotated data in our experiments comes from
the WSD evaluation framework of \cite{raganato2017},
which consists of five datasets
from five shared tasks on WSD:
Senseval-2 (SE2),
Senseval-3 (SE3),
SemEval-07 (S07),
SemEval-13 (S13),
and
SemEval-15 (S15).
The data is annotated using the WordNet 3.0 sense inventory,
and POS tagged with a low-granularity tag set
(nouns, verbs, adjectives, and adverbs).
We treat words that differ only in their part of speech as distinct word types.

We designate the oldest set (SE2) as our development set,
on which we tune the following parameters:
the number of companion words, which we set to $k = 20$,
and 
the linear weights defined in Section~\ref{combine},
which we set to $(\chi_1, \chi_2, \chi_3) = (0.5, 0.4, 0.1)$.

Our text corpus is
the OpenSubtitles2018 English-French bitext \cite{lison2016},
which contains roughly 42M sentences from various domains.
For consistency, 
we extract companions
and induce all types of vectors from the English side of the 
corpus.\footnote{Note that we could use any bitext for this purpose,
and that 
all vectors except the MFT vectors
could also be derived from other monolingual corpora.}
We use \textsc{word2vec}
to compute 200-dimensional 
vector embeddings
for the English and French vocabularies independently,
using the skip-gram model.
All other parameters are set to their default values.
To identify the MFT of each word,
we compute a bi-directional word alignment
of the bitext using GIZA++ \cite{och2003} 
with the default parameter settings. 
While our method is language-independent,
we perform the evaluation on English
because of the availability of a large sense-annotated corpus (SemCor).

\subsection{{\bve}}
\label{systems}

For the purpose of comparison, 
we re-implemented 
the {\bve} system \cite{bhingardive2015},
which, to the best of our knowledge,
is the most recent system to specifically consider the task of \mfsTask{}.
To make the comparison fair,
we use the same English word vectors
with \bve{}
as with our method.
We validated our reimplementation
by replicating the results on the noun subsets of the SE2 and SE3 datasets,
which are reported in the original paper.
While the replication experiment was conducted on nouns only,
we test on all parts of speech in the remainder of this paper. 

\subsection{Intrinsic Evaluation}
\label{ineval}

Our principal evaluation experiment is 
a direct intrinsic evaluation of {\mfsTask} systems:
how frequently does each system correctly identify the MFS of the given target?
This kind of an evaluation is only possible for English,
thanks to the availability of SemCor \cite{miller1993},
which is a relatively large sense-annotated corpus.\footnote{For consistency,
we use the version of SemCor made available by \cite{raganato2017}.}
We assume that the MFS for each word type 
is the sense that occurs with the highest frequency in SemCor.
We compute the accuracy of each method as
the proportion of word types for which its prediction matches the MFS.
If two or more senses are tied for the highest count,
we consider any of them to be a correct prediction.

SemCor is a resource unique to English,
composed of 226,034 sense-annotated word tokens,
which represent 22,436 word types.  
51.6\% of word types are polysemous,
that is, they occur in multiple WordNet synsets.
A randomly selected sense has a 67.6\% chance of being the MFS.
It should be noted that most of the word types are relatively infrequent, 
with the average count of 11, and the median count of 2.
The average number of word tokens per sense is 6.8,
and the corresponding median is again 2.

\begin{table}[t]
\centering
\begin{tabular}{|c||c|c|}
\hline
System     & All Words       &  Noun Sample \\\hline\hline 
{\Random}  &         67.6    &     26.0  \\\hline
\bve       &         73.9    &     48.0  \\\hline
\ourMethod &         75.2    &     48.8  \\\hline
{\wts}     & \textbf{77.9}   & \bf{58.5} \\\hline
\hline
PN18 (EnDi)    &        71.4     &     47.4      \\\hline 
\hline
MKWC04     & n/a             &     54    \\\hline
\end{tabular}
\caption{Intrinsic evaluation results on the {\mfsTask} task on SemCor
(in \% accuracy).}
\label{results-in}
\end{table}

The evaluation results are shown in Table~\ref{results-in}.
Both methods proposed in this paper outperform {\bve}.
Our WordNet-based {\wts} method
has a higher \mfsTask{} accuracy than either of the
methods based on word embeddings, \bve{} and \ourMethod{}.
This result 
confirms the utility of the concept of word {\relatives},
and provides strong support to our hypothesis that
they
convey a strong signal about the word's MFS.

To make our comparison more robust,
we replicate, as closely as possible,
the intrinsic evaluation performed by \cite{mccarthy2004},
henceforth referred to for brevity as ``MKWC04''.
To this end, 
this evaluation is performed on a sample of the words in SemCor;
specifically,
evaluation is performed only on polysemous nouns
(with respect to WordNet 3.0)
which occur at least three times in SemCor
and which have a single MFS (i.e. no ties).
This is slightly different from the MKWC04 evaluation,
which uses an older version of WordNet (1.6)
and a correspondingly older version of SemCor,
and which further samples the nouns to be evaluated on
by choosing nouns for which a certain amount of grammatical information
is available in a parsed corpus which they use as a resource.
Since these differences do not make the task easier,
it is fair to compare our results to what MKWC04 report.

The results of this ``noun sample'' intrinsic evaluation on SemCor
are also reported in Table \ref{results-in}.
Again, the random selection baseline is outperformed by all other systems.
Our \wts{} method clearly outperforms the reported result
of \cite{mccarthy2004}.
Interestingly, both \wts{} and MKWC04 outperform the vector-based methods.
This highlights the particularly strong performance 
of the $jcn$ similarity measure
on nouns as compared to other parts of speech.

Finally, we compare to the EnDi SDL method of \cite{pasini2018},
henceforth referred to as ``PN18''.
using the output data made available by the authors,
both on the full vocabulary,
and on the aforementioned noun sample.
For each word in the output,
we take the MFS to be the sense 
that is assigned the highest probability by EnDi.
To account for 8,434 SemCor words that are not in the PN18 output data,
we apply a random-selection backoff,
which yields the average accuracy of 84.4\% on those words.
The results show that EnDi is behind the other tested methods,
but it must be stressed that 
the task of sense distribution learning is more general than \mfsTask{}.

\subsection{WSD Evaluation}
\label{exeval}

\begin{table}[t]
  \centering
    \begin{tabular}{|c||c||c|c|c|c||c|}
    \hline
    System             & SE2 & SE3       & S07       & S13       & S15      & ALL      \\\hline\hline
    {\bve}             &     54.8  &     52.0  &     38.2  & \bf{55.2} &     54.5 &     53.1 \\\hline
    \ourMethod         & \bf{56.4} & \bf{53.8} & \bf{40.6} &     54.9  &     54.0 & \bf{54.1}\\\hline
    {\wts}             &     51.5  &     47.0  &     37.5  &     54.1  & \bf{55.0}&     50.7 \\\hline
    \hline
    {\Oracle}          &     65.6  &     66.0  &     54.5  &     63.8  &     67.1 &     65.5  \\\hline
    \hline
    \lext              &     50.6  &     44.5  &     32.0  &     53.6
&     51.0 & n/a
\\\hline
  \end{tabular}
  \caption{Extrinsic evaluation of the {\mfsTask} systems on the WSD task
  (in \% $F_1$-score).
  }
  \label{results-ex}
\end{table}

An indirect, extrinsic evaluation of an {\mfsTask} system is
to apply it to the WSD task 
by simply predicting the most frequent sense for each word,
regardless of its context.
Unlike the intrinsic evaluation above,
WSD evaluation is conducted on the level on word tokens, rather than types,
with multiple instances of the same word
contributing independently to the results.

In addition to the systems evaluated in the previous section,
we also include the results of two other methods,
as reported by \cite{raganato2017,raganato2017emnlp}.
\emph{\Oracle}
outputs the most frequent sense according to the SemCor annotations,
demonstrating what a ``perfect'' \mfsTask{} system
(applied to SemCor) could achieve.
\emph{\lext} \cite{banerjee2003}
is a WordNet-based extension of the classic Lesk algorithm \cite{lesk1986},
and serves as a strong unsupervised baseline WSD system.
Unlike the {\mfsTask} systems,
it disambiguates words at the level of individual tokens,
rather than word types,
according to the context of each instance.

The results are reported in Table~\ref{results-ex}.
\ourMethod{} is the top-performing unsupervised method on 
the development set (SE2),
two of the other data sets (SE3 and S07),
and is within 1\% of the best result on the other two data sets (S13 and S15).
Following 
recent work \cite{raganato2017emnlp},
we also test on the concatenation of the five datasets,
on which \ourMethod{} also obtains the best result. 

Another interesting observation is 
that all three MFS-based approaches 
outperform the \lext{} method.
Unlike \mfsTask{} methods 
\lext{} has the ability to 
select different senses for different tokens of the same type
depending on context,
These results,
together with the high {\oracle} ceilings shown in Table~\ref{results-ex},
confirm the importance of accurate
{\mfsTask} for the WSD task.

Finally, a key difference from the intrinsic evaluation results 
reported in Section \ref{ineval},
is the superior performance of 
our vector-based \ourMethod{} method
versus the WordNet-based \wts{} method.
These results show that
an \mfsTask{} system which is strictly better at 
detecting the most frequent sense of a word
may not produce the best results when its output is used for WSD.
This also demonstrates that both of the systems 
which we have developed in this paper have merit, 
depending on the proposed application:
\wts{} gives better type-level accuracy 
when the MFS itself is desired,
while \ourMethod{} provides better token-level results 
when the goal is to apply the output to perform WSD.

\subsection{Ablation Experiments}
\label{ablation}

\begin{table}[t]
\centering
\begin{tabular}{|c|c|c|}
\hline
                        & Intrinsic & Extrinsic \\
\ourMethod{} Variant    & (SemCor)  & (SE2) \\\hline\hline

Full system    & \bf{75.2} & \bf{56.4} \\\hline \hline

Word vector         & 74.5 & 55.2 \\\hline 
\Relatives{} vector & 67.4 & 53.2 \\\hline 
MFT vector          & 71.9 & 49.8 \\\hline 
\hline

Knowledge-light & 71.9 & 52.7 \\ 
\hline
\end{tabular}
\caption{
Ablation results
for the intrinsic (in \% accuracy) 
and extrinsic experiments (in \% $F_1$-score).
}
\label{results-ab}
\end{table}

\ourMethod{} has a highly modular structure,
making it adaptable to a variety of alternative settings.
In this section, we perform a series of ablation experiments,
in which various components or sources of information
are removed from \ourMethod{}
to measure their impact.
We perform intrinsic evaluation experiments on SemCor (all words),
and extrinsic evaluation experiments on the SE2 WSD dataset.

Our first set of ablation experiments evaluate
the utility of the three sources of information used by {\ourMethod}:
the vector of the target word,
the average of the vectors of its {\relatives},
and the transformed vector of its most frequent translation.
We run three \emph{feature ablation} experiments,
each using only one of these vectors,
with the $\chi$ coefficients corresponding to the other vectors
set to 0.

The results of this experiment in Table \ref{results-ab}
show that,
if only one of these three vectors can be constructed,
sense vectors are best compared to word vectors,
as using only this vector gives better results than using only one
of the other vectors.
For intrinsic evaluation, using only the MFT vector gives better results
than using only the \relatives{} vector;
on the extrinsic evaluation, the reverse is true.
This once again shows the potential for disagreement
between intrinsic and extrinsic evaluations of \mfsTask{} systems.

In order to measure the impact of WordNet
as a source of linguistic knowledge,
we also perform a \emph{knowledge ablation} experiment,
in which \ourMethod{}
only has access to WordNet's glosses and examples,
and not to the WordNet synset hierarchy 
or any of its information on semantic relationships
such as synonymy or hypernymy.
This decreases the number of keywords 
that are available for the construction of sense vectors,
and essentially reduces WordNet to a machine-readable dictionary. 
This setting,
which we refer to as {\em knowledge-light},
emulates the circumstances of 
working with less well-studied languages,
for which machine-readable dictionaries are available,
but WordNet-like knowledge bases are not.
In this setting,
our parameter tuning procedure for \ourMethod{} 
yields the values of $(\chi_1, \chi_2, \chi_3) = (0.4, 0.1, 0.5)$ 
on the development set,
which suggests that
our innovation of leveraging a parallel corpus
helps recover some linguistic information lost in this setting.

Table \ref{results-ab} shows that
the decline in the performance of \ourMethod{} in the knowledge-light setting
compared to the standard 
knowledge-based version
is relatively small, with a drop in accuracy of only 3.3\%
on \mfsTask{} on SemCor.
This shows that \ourMethod{} ultimately has lower information requirements
compared to prior work,
and is applicable for low-resource settings.

WordNet glosses may include usage examples for a given sense.
As a final ablation experiment, 
we measure the effect of removing the access to these examples
on the tested methods.
We find that this lowers the results in all cases;
however, the magnitude of the effect varies across the three systems.
In an extrinsic evaluation on the SE2 WSD dataset,
we observe
a decrease of only 0.5\% F1 for \ourMethod{}
in the standard knowledge-based setting, 
but 2.9\% in the knowledge-light setting,
while \bve{} drops by 1.3\%.
A pattern is apparent in these results: 
the more linguistic knowledge a system uses, 
the less it benefits from the inclusion of examples. 

\subsection{Error Analysis}
\label{erranal}

An example illustrating the advantage of using translations 
involves the noun {\em brow},
which has three principal senses:
``hair above the eye'' (MFS),
``part of the face'',
and ``peak of a hill.''
The {\wts} approach incorrectly selects the last sense,
which is actually the \emph{least} frequent sense.
\ourMethod{} is able to identify the MFS
by leveraging the fact that each of the senses
translates into a different French word
({\em sourcil}, {\em front}, and {\em sommet}, respectively).

The difficulty of the task is illustrated by the verb {\em bow}.
The MFS is ``bend one's knee or body, or lower one's head'',
but WordNet contains also two other similar senses:
``bend the head or the upper part of the
body in a gesture of respect or greeting'',
and ``bend one's back forward from the waist on down.''
The challenge of distinguishing between these senses
is highlighted by the fact that all three sense include
an almost identical usage example:
``she/he bowed before the king/queen". 
We conclude that,
although WordNet is a standard evaluation resource,
its fine-grained sense inventory may not be optimal for the WSD task.
This is in accordance with prior work which has shown that WSD performance 
improves when performed with respect to 
less granular sense inventories \cite{navigli2006}.

Although our knowledge-light version commits errors 
on a number of instances 
for which the WordNet information is crucial,
there are also hundreds of words where it outdoes all other tested methods.
For example, 
it correctly identifies the MFS of the verb {\em bore}
as ``cause to be bored.''
The fact that the other methods choose instead
the sense of ``make a hole, especially with a pointed power or hand tool''
can be attributed to the fact that
this sense has a more detailed gloss,
and is accompanied by 4 usage examples,
as opposed to no examples for the MFS.
It seems that the varying amounts of extra information 
available in WordNet for different senses
may be a source of confusion for the knowledge-intensive methods.

We hope that our findings will motivate further research into knowledge-light
{\mfsTask} and WSD. Indeed, the trend in recent years has been
to augment WSD systems with increasingly rich and complex
sources of linguistic knowledge, such as BabelNet \cite{navigli2012}.
That systems with more resources available perform better in general 
is unsurprising,
and raises the question of whether recent advances in WSD are primarily
due to the addition of new sources of linguistic knowledge,
rather than algorithmic innovation.
Development and comparison of systems in resource-controlled settings
could provide useful insights into 
how both WSD and \mfsTask{} could be improved.

\section{Conclusion}
\label{conclusion}

We have presented two novel {\mfsTask} methods,
one which uses a sense-to-sense similarity measure
to find the sense which is most related to the \relatives{} of the target,
and another 
which uses cross-lingual vector representations of words
derived using a bitext.
Our intrinsic and extrinsic evaluations show that the two methods
perform well in comparison to previous work,
and that they outperform a strong knowledge-based baseline (extended Lesk)
when applied to word sense disambiguation.
The ablation experiments
demonstrate that our innovation of leveraging a bitext
helps recover some of the lost information, improving results.
In short, we have established that our contributions
of defining and applying the \relatives{}
and most frequent translations
lead to improved performance in \mfsTask{}.

In the future, we plan to explore 
ways of applying these concepts directly to word sense disambiguation,
as well as
leveraging WordNet-based similarity measures in a cross-lingual setting,
and enhancing such measures with word vectors.

\section*{Acknowledgements}

We thank Tommaso Pasini for the assistance
with the SDL output data.

This research was supported by 
the Natural Sciences and Engineering Research Council of Canada, 
Alberta Innovates, and Alberta Advanced Education.

\bibliography{wsd}
\bibliographystyle{IEEEtran.bst}
\balance
\end{document}